\documentclass[conference]{IEEEtran}
\usepackage{amsmath,amssymb,amsfonts}
\usepackage{algorithm}

\usepackage{algpseudocode}
\usepackage{graphicx}
\usepackage{tabularx}
\usepackage{array}
\usepackage{cite}
\usepackage{stfloats}
\usepackage{textcomp}
\usepackage{xcolor}
\def\BibTeX{{\rm B\kern-.05em{\sc i\kern-.025em b}\kern-.08em
    T\kern-.1667em\lower.7ex\hbox{E}\kern-.125emX}}

\begin{document}

\title{A Special Point Skeleton Reconstruction Algorithm for Dynamic Multiobjective Optimization
\\

}

\author{\IEEEauthorblockN{1\textsuperscript{st}  GuangXian Gan}
\IEEEauthorblockA{\textit{South China Normal University}\\
GuangZhou, China \\
2025023287@m.scnu.edu.cn
}
\and
\IEEEauthorblockN{2\textsuperscript{nd} MinRong Chen}
\IEEEauthorblockA{\textit{South China Normal University} \\
GuangZhou, China \\
chenminrong@scnu.edu.cn}

}

\maketitle
\begin{abstract}
To address the issue that existing dynamic multi-objective optimization algorithms mainly rely on individual migration or independent special point sampling after environmental changes, while failing to fully exploit the structural relationships among representative solutions, a Special Point Skeleton Reconstruction based Dynamic Multi-Objective Evolutionary Algorithm (SPSR-DMOEA) is proposed. First, the centroid, knee points, and extreme points are extracted from the Pareto optimal solution set of the current environment, and their positions in the new environment are adaptively predicted according to their movement velocities across consecutive environments. Subsequently, in the decision space, the centroid is connected with other anchor points, and a minimum spanning tree is constructed among the non-centroid anchor points, thereby establishing a prediction skeleton capable of describing the overall population structure. According to the lengths of the skeleton edges, the number of individuals allocated to each edge is determined proportionally. Candidate solutions are uniformly generated along each edge, and random orthogonal perturbations are introduced to expand the search region around the skeleton. Experimental results on the DF dynamic multi-objective benchmark suite demonstrate the effectiveness of the proposed method in dynamic tracking capability.
\end{abstract}


\section{Introduction}
Many real-world optimization problems involve multiple conflicting objectives and are often affected by continuously changing environments. Different from static multi-objective optimization problems, dynamic multi-objective optimization (DMO) problems require algorithms to effectively track the changing Pareto-optimal solutions and maintain decision-making performance after environmental variations. Such problems widely exist in various fields, including mineral processing, manufacturing, transportation, and financial management. For example, Ding et al. investigated dynamic evolutionary multi-objective optimization for raw ore allocation under changing resource and production conditions~\cite{8331273}. Dynamic multi-objective flexible job shop scheduling and urban rail transit scheduling have also been studied to address real-time changes in production and transportation environments~\cite{YUAN2025112787,SHEN2025101960}. Moreover, predictive multi-period multi-objective portfolio optimization has been developed to adapt investment strategies to dynamic market conditions~\cite{ABOLMAKAREM2023109450}. These practical applications highlight the importance of developing effective algorithms for solving dynamic multi-objective optimization problems.

Prediction-based approaches have become an important research direction in DMOEAs due to their ability to utilize historical evolutionary information to estimate future environments. Zhou et al. proposed a population prediction strategy that employed the evolutionary trajectories of historical populations to predict the distribution of future populations, providing an effective way to accelerate the adaptation process after environmental changes~\cite{6471286}. Subsequently, transfer learning techniques were introduced into dynamic multi-objective optimization to exploit knowledge from previous environments and improve the prediction accuracy of new populations~\cite{8100935}. Recently, some studies have further considered the structural characteristics of Pareto-optimal populations. For example, Huang et al. developed a knee-guided prediction model that utilized knee points to guide population re-initialization and improve the prediction performance~\cite{HUANG2026102358}. However, most existing prediction-based methods mainly focus on predicting individual solutions or generating new populations through solution migration, while the intrinsic structural relationships among representative solutions within the Pareto-optimal population remain insufficiently explored.

To address the aforementioned limitations, this paper proposes a special point skeleton reconstruction based dynamic multi-objective evolutionary algorithm (SPSR-DMOEA). Instead of directly predicting individual solutions, the proposed method focuses on predicting several representative points that characterize the structural information of the Pareto-optimal population, including the centroid, knee point, and extreme points. By establishing a skeleton structure among these representative points in the decision space, the proposed method reconstructs the population after environmental changes and generates candidate solutions along the skeleton with adaptive perturbation. In this way, the predicted population can preserve both the global distribution and structural characteristics of the original Pareto-optimal population. The main contributions of this work are summarized as follows:

A special point skeleton Reconstruction strategy is proposed. Using the skeleton constructed by the predicted knee point, centroid, and extreme points to reconstruct the population.

The remainder of this paper is organized as follows. Section II introduces the formulation of dynamic multi-objective optimization problems, reviews prediction-based dynamic multi-objective evolutionary algorithms, and presents the concept of knee points. Section III details the proposed special point skeleton reconstruction framework. Section IV reports the experimental results and discusses the findings. Finally, Section V concludes the paper and outlines future research directions.

\section{RELATED WORK}
\subsection{Dynamic Multi-objective Optimization Problems}
Considering a minimization problem, a dynamic multi-objective optimization problem (DMOP) can be mathematically described as follows~\cite{10611891}:
\begin{equation}
\begin{cases}
\text{minimize} \quad F(\mathbf{x}, t)=\{f_1(\mathbf{x}, t), f_2(\mathbf{x}, t), \ldots, f_m(\mathbf{x}, t)\} \\
\text{subject to} \quad \mathbf{x} \in \Omega
\end{cases}
\end{equation}
where \(\mathbf{x}=(x_1,x_2,\ldots,x_d)\) represents a \(d\)-dimensional decision vector, \(\Omega\) denotes the decision space, \(m\) indicates the number of objectives, and \(t\) is the time index associated with environmental changes. The objective vector \(F(\mathbf{x},t)\) contains \(m\) objective functions that vary with the dynamic environment.

\textit{Definition 1 (Pareto Dominance):} For two solutions \(\mathbf{x}_a\) and \(\mathbf{x}_b\) in \(\Omega\) at time \(t\), \(\mathbf{x}_a\) is said to dominate \(\mathbf{x}_b\), denoted as \(\mathbf{x}_a \prec_t \mathbf{x}_b\), if the following conditions are satisfied:
\begin{equation}
\begin{cases}
f_i(\mathbf{x}_a,t) \leq f_i(\mathbf{x}_b,t), & \forall i \in \{1,2,\ldots,m\}, \\
f_j(\mathbf{x}_a,t) < f_j(\mathbf{x}_b,t), & \exists j \in \{1,2,\ldots,m\}.
\end{cases}
\end{equation}

\textit{Definition 2 (Pareto-Optimal Set):} A solution \(\mathbf{x}^* \in \Omega\) is Pareto-optimal at time \(t\) if no other solution in \(\Omega\) can dominate it. Therefore, the Pareto-optimal set (POS) is defined as the set of all non-dominated solutions:
\begin{equation}
POS(t)=\left\{\mathbf{x}^* \in \Omega \mid \nexists \mathbf{x} \in \Omega,\ \mathbf{x} \prec_t \mathbf{x}^* \right\}.
\end{equation}

\textit{Definition 3 (Pareto-Optimal Front):} The Pareto-optimal front (POF) represents the mapping of the Pareto-optimal set from the decision space to the objective space. It can be expressed as:
\begin{equation}
POF(t)=\left\{F(\mathbf{x}^*,t) \mid \mathbf{x}^* \in POS(t) \right\}.
\end{equation}

\subsection{Predict-based DMOEA}
Prediction-based approaches have become an important research direction in dynamic multi-objective evolutionary optimization because they can exploit historical evolutionary information to estimate future population states and improve the adaptation efficiency after environmental changes. The main idea of these methods is to predict the movement tendency or distribution characteristics of Pareto-optimal solutions, thereby guiding the generation of populations in subsequent environments.

To improve the accuracy of population prediction, various strategies have been developed by exploiting different types of evolutionary information. Sun et al. proposed a quantile-guided dual prediction strategy, which utilized quantile information to capture diverse evolutionary trends and improve prediction robustness under dynamic environments~\cite{SUN2021751}. Feng et al. introduced a guided prediction strategy based on regional multi-directional information fusion, where evolutionary information from different regions was integrated to enhance prediction performance~\cite{FENG2024120565}. Xu et al. further proposed a cluster prediction strategy with induced mutation, which utilized clustering information to characterize population distribution and generate adaptive solutions after environmental changes~\cite{XU2024120193}. In addition, Zhang et al. developed an inverse Gaussian process modeling approach that constructed a prediction model in the objective space to estimate the future Pareto-optimal solutions~\cite{9440867}.

Besides direct prediction models, knowledge transfer mechanisms have been widely incorporated into DMOEAs to reuse valuable information from previous environments. Jiang et al. proposed a fast dynamic evolutionary multi-objective algorithm based on manifold transfer learning, which transferred low-dimensional manifold information extracted from historical populations to accelerate adaptation in new environments~\cite{9097186}. Lin et al. introduced a knowledge transfer and maintenance framework that preserved useful historical knowledge and adaptively transferred it to future environments~\cite{10292939}. Zou et al. proposed a knowledge transfer method based on mixture models, where probabilistic models were employed to describe and transfer evolutionary knowledge under dynamic changes~\cite{10982147}. Hu et al. further considered environmental similarity identification and knowledge transfer, enabling the algorithm to select appropriate historical information for different dynamic environments~\cite{11122890}.

Apart from prediction and knowledge transfer strategies, other adaptive response mechanisms have also been investigated to enhance the robustness of DMOEAs. Yu et al. proposed a dual-space detection based adaptive response algorithm that integrated decision-space and objective-space information to improve environmental adaptability~\cite{YU2025102092}. Liu et al. developed a dual mutation-based evolutionary algorithm for dynamic multi-objective optimization with undetectable changes, which enhanced population diversity and adaptation ability through a dual mutation mechanism~\cite{10587214}.

Although existing prediction-based DMOEAs have achieved significant progress, most methods mainly focus on estimating future population distributions, transferring historical knowledge, or generating new individuals through migration and variation operations. The intrinsic geometric relationships among representative solutions in Pareto-optimal populations are rarely considered. Therefore, utilizing the structural information contained in representative points to reconstruct population distributions provides a promising direction for dynamic multi-objective optimization.
\subsection{knee point}
The concept of knee points was introduced by Branke et al.~\cite{10.1007/978-3-540-30217-9_73} to identify representative solutions with significant trade-off characteristics on the Pareto-optimal front. Specifically, a knee point corresponds to a region where improving one objective would lead to a considerable degradation of other objectives. Due to its ability to reflect important distribution characteristics of Pareto-optimal solutions, knee points are widely used as representative points in multi-objective optimization.

\section{Proposed Method}

\subsection{Overall Framework}

The overall workflow of SPSR-DMOEA is as follows: When no environmental change is detected, the population is evolved using a static multi-objective optimization algorithm. Once an environmental change is detected, the population is randomly reinitialized for the first change; for subsequent changes, the corresponding response mechanism is activated. The algorithm predicts the centroid, knee points, and extreme points of the population at the next time step, and utilizes the decision-space skeleton constructed by these points to generate the new population.

\subsection{Special Point Skeleton-Based Reconstruction Prediction}
To address environmental changes, we predict the centroid, knee points, and extreme points of the next environment, and construct a generation skeleton for the predicted population based on these special points. The centroid is calculated as follows:
\begin{equation}
\mathbf{c}^{t}
=
\frac{1}{N}
\sum_{i=1}^{N}\mathbf{x}_{i}^{t}
\label{eq:centroid_calculate}
\end{equation}
where $\mathbf{c}^{t}$ denotes the centroid of the POS in environment $t$,
$N$ is the number of solutions in the current POS, and
$\mathbf{x}_{i}^{t}$ represents the decision vector of the $i$-th solution
in environment $t$.

When predicting the special points in the next environment, we adopt the method proposed in \cite{ZHANG2025113072}, which employs a linear extrapolation with an adaptive factor considering the degree of environmental change. The calculation formula is given as follows:

\begin{equation}
\mathbf{X}^{t+1}
=
\mathbf{X}^{t}
+
\boldsymbol{\alpha}^{t}
(\mathbf{X}^{t}-\mathbf{X}^{t-1})
\label{eq:feature_prediction}
\end{equation}
where $\mathbf{X}^{t+1}$ denotes the predicted position of a special point
in environment $t+1$, while $\mathbf{X}^{t}$ and $\mathbf{X}^{t-1}$
denote the positions of the corresponding special point in environments
$t$ and $t-1$, respectively. Knee points in consecutive environments are matched to their nearest counterparts in the objective space. $\boldsymbol{\alpha}^{t}$ is the adaptive prediction factor used to adjust the prediction step size.

\begin{equation}
\alpha^{t}
=
\frac{
\|\mathbf{X}^{t}-\mathbf{X}^{t-1}\|
}{
\|\mathbf{X}^{t-1}-\mathbf{X}^{t-2}\|
}
\label{eq:adaptive_factor}
\end{equation}
where the notation $\|\cdot\|$ denotes the Euclidean norm.A larger $\alpha^{t}$ indicates a more significant change, suggesting a larger prediction step.

We use the predicted special points to construct the skeleton. Specifically, the centroid is connected to all knee points, and a minimum spanning tree is constructed among the knee points. These special points and their corresponding edge relationships constitute the skeleton used for subsequent population reconstruction.

\begin{algorithm}
\caption{Feature Prediction and Skeleton Construction}
\label{alg:skeleton_prediction}
\begin{algorithmic}[1]
\Require Historical POS sets $\mathcal{H}$, population size $N$
\Ensure Predicted population $P^{t+1}$

\State calculate  centroid $\mathbf{c}^{t}$ using Eq.~(\ref{eq:centroid_calculate}) 
\State calculate extreme points $\mathbf{E}^{t}$ and knee points $\mathbf{K}^{t}$

\State Predict centroid $\mathbf{c}_{pre}^{t+1}$, extreme points $\mathbf{E}_{pre}^{t+1}$ and knee points $\mathbf{K}_{pre}^{t+1}$
using Eq.~(\ref{eq:feature_prediction})

\State Construct the anchor set
$\mathbf{A}=
[\mathbf{c}_{pre}^{t+1};
\mathbf{E}_{pre}^{t+1};
\mathbf{K}_{pre}^{t+1}]$

\State Normalize $\mathbf{A}$ and remove duplicate anchors

\State Connect the centroid to all non-centroid anchors

\State Construct an MST among the non-centroid anchors

\State Combine the above edges to obtain the predicted skeleton
$\mathcal{G}=(\mathbf{A},\mathcal{E})$

\State Generate $P^{t+1}$ from $\mathcal{G}$
using Algorithm~\ref{alg:skeleton_reconstruction}

\State \Return $P^{t+1}$
\end{algorithmic}
\end{algorithm}

\subsection{Population Reconstruction}
Based on the predicted population skeleton, the number of individuals generated on each edge is allocated according to the edge length. The allocation formula is defined as follows:
\begin{equation}
q_e
=
\left(N-|\mathbf{A}|\right)
\frac{L_e}
{\displaystyle\sum_{h\in\mathcal{E}}L_h},
\label{eq:edge_allocation}
\end{equation}
where $L_e$ denotes the length of edge $e$, and $q_e$ is the real-valued
number of individuals assigned to edge $e$. Therefore, a longer skeleton edge is
allocated more individuals to provide sufficient coverage of the
corresponding region.

When generating new individuals, we first uniformly generate basic individuals along each edge, and then add a random perturbation orthogonal to the corresponding edge to each basic individual. The calculation formula is given as follows:

\begin{equation}
\mathbf{x}_{e,s}
=
(1-\frac{s}{q_e+1})\mathbf{a}_i
+
\frac{s}{q_e+1}\mathbf{a}_j
+
\mathbf{u}_{e}^{\perp}
\qquad
\label{eq:individual_generation}
\end{equation}
where $\mathbf{x}_{e,s}$ denotes the $s$-th individual generated on
edge $e$ and $\mathbf{u}_{e}^{\perp}$ is a random perturbation vector in the orthogonal direction with respect to the edge $e$.
\begin{algorithm}
\caption{Population Reconstruction Based on the Predicted Skeleton}
\label{alg:skeleton_reconstruction}
\begin{algorithmic}[1]
\Require Normalized predicted skeleton
$\mathcal{G}=(\mathbf{A},\mathcal{E})$,
population size $N$
\Ensure Predicted population $P^{t+1}$

\State Initialize $\widetilde{P}^{t+1}\gets\mathbf{A}$

\State Allocate the remaining $N-|\mathbf{A}|$ individuals
to the skeleton edges using Eq.~(\ref{eq:edge_allocation})

\State Adjust the allocation using the largest-remainder method

\ForAll{edge $e=(i,j)\in\mathcal{E}$}
    \For{$s=1$ to $n_e$}
        \State Generate a random direction
        $\mathbf{u}_{e}^{\perp}$ 

        \State Generate $\mathbf{x}_{e,s}$
        using Eq.~(\ref{eq:individual_generation})

        \State Add $\mathbf{x}_{e,s}$ to $\widetilde{P}^{t+1}$
    \EndFor
\EndFor

\State Map $\widetilde{P}^{t+1}$ to the original decision space

\State Randomly replace $10\%$ of $P^{t+1}$
with uniformly generated individuals

\State \Return $P^{t+1}$
\end{algorithmic}
\end{algorithm}
\section{RESULTS AND ANALYSIS}
\subsection{Test Problems and Performance Indicator}

To assess the effectiveness of the proposed SPSR-DMOEA, the DF benchmark suite developed for the CEC2018 dynamic multi-objective optimization competition is employed \cite{Jiang2018BenchmarkPF}. All experiments are conducted on the PlatEMO platform \cite{8065138}. The DF suite consists of various dynamic multi-objective test problems with diverse characteristics, including time-varying Pareto sets, changing Pareto fronts, irregular Pareto-front geometries, and disconnected Pareto fronts. These features make the DF benchmark suite a suitable platform for comprehensively evaluating the convergence, diversity preservation, and environmental tracking capability of dynamic multi-objective evolutionary algorithms.

In this study, the inverted generational distance (IGD) and the mean
inverted generational distance (MIGD) are used as performance
indicators. At time $t$, IGD is defined as
\begin{equation}
\mathrm{IGD}(PF_t^{*}, PF_t^{e}) =
\frac{\sum_{\mathbf{p} \in PF_t^{*}} d(\mathbf{p}, PF_t^{e})}
{|PF_t^{*}|}
\end{equation}
where $PF_t^{*}$ represents the set of reference points uniformly sampled from the actual Pareto front at time $t$, $PF_t^{e}$ represents the Pareto front approximation obtained by the proposed algorithm, and $d(\mathbf{p}, PF_t^{e})$ indicates the shortest Euclidean distance between the reference point $\mathbf{p}$ and the solutions in $PF_t^{e}$.

To evaluate the overall performance over all dynamic environments,
MIGD is calculated by averaging IGD values over all time instances:
\begin{equation}
\mathrm{MIGD} =
\frac{\sum_{t \in T} \mathrm{IGD}(PF_t^{*}, PF_t^{e})}
{|T|}
\end{equation}
where $T$ denotes the set of discrete time steps during a single run, and $|T|$ represents the total number of environmental changes. A lower MIGD value indicates superior overall optimization performance.

\subsection{Comparing Algorithms and Parameter Settings}

To verify the effectiveness of the proposed SPSR-DMOEA, four
representative dynamic multi-objective evolutionary algorithms are
selected as comparison algorithms, including SVR-DMOEA
\cite{8790005}, KGB-DMOEA \cite{9953975}, AE-DMOEA
\cite{9210737}, and DIP-DMOEA \cite{10508204}. For a fair comparison,
all algorithms adopt NSGA-II as the static optimizer, and the
population size is set to 100 for all test problems.Moreover, all algorithms are evaluated under the same dynamic setting with a total budget of 30000 function evaluations.

The dynamic characteristics of the test problems are controlled by the
severity parameter $n_t$ and the frequency parameter $\tau_t$. In this
study, three different dynamic settings are considered, namely
$(n_t,\tau_t)=(5,10)$, $(10,10)$, and $(20,10)$.For the proposed SPSR-DMOEA, the number of the knee points is set to 10.

\subsection{Results and Analysis}
\begin{table*}[!t]
\caption{Mean and Standard Deviations of MIGD Results of SPSR-DMOEA and Competitors on the DF Test Suite}
\label{tab:migd}
\centering
\footnotesize
\setlength{\tabcolsep}{2.4pt}
\renewcommand{\arraystretch}{1.08}
\begin{tabularx}{\textwidth}{c c *{5}{>{\centering\arraybackslash}X}}
\hline
\textbf{Problem} & \textbf{$(n_t,\tau_t)$} & \textbf{AE-DMOEA} & \textbf{KGB-DMOEA} & \textbf{DIP-DMOEA} & \textbf{SVR-DMOEA} & \textbf{SPSR-DMOEA} \\
\hline
DF1 & $(5,10)$ & 9.3056e-2 (5.03e-3) $-$ & 8.1585e-2 (7.92e-3) $-$ & 8.5227e-2 (1.16e-2) $-$ & 2.3107e-1 (6.72e-2) $-$ & \textbf{6.1212e-2 (6.36e-3)} \\
  & $(10,10)$ & 3.8083e-2 (2.24e-3) $-$ & 7.3369e-2 (5.75e-3) $-$ & 5.7518e-2 (1.22e-2) $-$ & 8.1731e-2 (2.75e-2) $-$ & \textbf{2.5983e-2 (2.93e-3)} \\
  & $(20,10)$ & 1.7815e-2 (2.67e-3) $=$ & 9.0201e-2 (4.09e-3) $-$ & 2.5138e-2 (4.13e-3) $-$ & 3.5138e-2 (5.00e-3) $-$ & \textbf{1.6406e-2 (1.62e-3)} \\
\hline
DF2 & $(5,10)$ & 1.3490e-1 (1.24e-2) $-$ & 7.6002e-2 (1.07e-2) $-$ & 8.1936e-2 (1.19e-2) $-$ & 1.9692e-1 (4.53e-2) $-$ & \textbf{6.3402e-2 (6.93e-3)} \\
  & $(10,10)$ & 1.4721e-1 (1.67e-2) $-$ & 7.7831e-2 (2.09e-2) $-$ & 8.0612e-2 (1.24e-2) $-$ & 1.3306e-1 (2.86e-2) $-$ & \textbf{5.4884e-2 (8.45e-3)} \\
  & $(20,10)$ & 1.4033e-1 (1.78e-2) $-$ & 8.8400e-2 (2.29e-2) $-$ & 7.7767e-2 (1.71e-2) $-$ & 1.0071e-1 (1.01e-2) $-$ & \textbf{2.9713e-2 (5.61e-3)} \\
\hline
DF3 & $(5,10)$ & 3.9814e-1 (3.64e-2) $-$ & 6.0392e-1 (4.89e-2) $-$ & 5.4503e-1 (9.17e-2) $-$ & 4.1652e-1 (5.75e-2) $-$ & \textbf{3.2950e-1 (5.63e-2)} \\
  & $(10,10)$ & \textbf{2.8136e-1 (5.30e-2)} $=$ & 6.2443e-1 (3.98e-2) $-$ & 3.7595e-1 (8.60e-2) $-$ & 3.0724e-1 (3.52e-2) $=$ & 2.9366e-1 (4.78e-2) \\
  & $(20,10)$ & 3.4328e-1 (5.42e-2) $-$ & 6.3248e-1 (4.20e-2) $-$ & 3.4486e-1 (7.28e-2) $-$ & 3.9311e-1 (3.93e-2) $-$ & \textbf{2.5597e-1 (5.27e-2)} \\
\hline
DF4 & $(5,10)$ & 1.7575e-1 (2.39e-2) $-$ & 9.6278e-1 (3.82e-1) $-$ & 4.2474e-1 (2.33e-1) $-$ & 2.6314e-1 (3.81e-2) $-$ & \textbf{1.4218e-1 (1.32e-2)} \\
  & $(10,10)$ & 1.5145e-1 (1.82e-2) $-$ & 1.0320e+0 (5.09e-1) $-$ & 3.1694e-1 (1.67e-1) $-$ & 2.5924e-1 (2.75e-2) $-$ & \textbf{1.3226e-1 (1.61e-2)} \\
  & $(20,10)$ & 1.9241e-1 (2.59e-2) $-$ & 6.3197e-1 (1.71e-1) $-$ & 2.8684e-1 (1.05e-1) $-$ & 3.5747e-1 (3.50e-2) $-$ & \textbf{1.4931e-1 (1.21e-2)} \\
\hline
DF5 & $(5,10)$ & 1.5768e-1 (2.11e-2) $-$ & 2.4124e-1 (2.46e-2) $-$ & 1.4160e-1 (1.82e-2) $-$ & 1.8822e-1 (4.38e-2) $-$ & \textbf{4.0653e-2 (8.03e-3)} \\
  & $(10,10)$ & 5.0856e-2 (6.56e-3) $-$ & 3.1744e-1 (3.05e-2) $-$ & 1.3288e-1 (4.79e-2) $-$ & 1.0463e-1 (2.77e-2) $-$ & \textbf{2.4226e-2 (2.88e-3)} \\
  & $(20,10)$ & \textbf{1.6365e-2 (1.94e-3)} $+$ & 4.1134e-1 (3.22e-2) $-$ & 3.8088e-2 (2.61e-2) $-$ & 3.4394e-2 (1.44e-2) $-$ & 2.0620e-2 (2.55e-3) \\
\hline
DF6 & $(5,10)$ & 3.3253e+0 (3.23e-1) $-$ & 3.2373e+0 (2.28e-1) $-$ & 4.0867e+0 (4.26e-1) $-$ & 5.2675e+0 (8.79e-1) $-$ & \textbf{1.9808e+0 (5.86e-1)} \\
  & $(10,10)$ & 2.7493e+0 (3.02e-1) $-$ & 3.3312e+0 (2.97e-1) $-$ & 3.8794e+0 (4.95e-1) $-$ & 4.7156e+0 (2.49e+0) $-$ & \textbf{2.2114e+0 (4.46e-1)} \\
  & $(20,10)$ & 1.1963e+0 (1.72e-1) $-$ & 2.5983e+0 (1.84e-1) $-$ & 2.5004e+0 (4.71e-1) $-$ & 1.7351e+0 (4.94e-1) $-$ & \textbf{9.4598e-1 (2.31e-1)} \\
\hline
DF7 & $(5,10)$ & 3.0230e-1 (1.47e-1) $-$ & 3.1021e-1 (7.89e-2) $-$ & 6.1756e-1 (5.73e-2) $-$ & 6.0314e-1 (1.41e-1) $-$ & \textbf{1.3607e-1 (5.25e-2)} \\
  & $(10,10)$ & 2.1739e-1 (1.22e-1) $-$ & 3.0906e-1 (9.96e-2) $-$ & 6.1442e-1 (3.49e-2) $-$ & 2.9560e-1 (7.91e-2) $-$ & \textbf{1.0194e-1 (3.24e-2)} \\
  & $(20,10)$ & 1.6803e-1 (5.03e-2) $-$ & 3.4008e-1 (1.33e-1) $-$ & 3.1109e-1 (5.78e-2) $-$ & 2.0561e-1 (2.10e-2) $-$ & \textbf{1.0615e-1 (1.81e-2)} \\
\hline
DF8 & $(5,10)$ & 2.4409e-2 (3.75e-3) $=$ & 1.2926e-1 (8.77e-3) $-$ & 3.0313e-2 (7.48e-3) $-$ & 2.9795e-2 (7.05e-3) $-$ & \textbf{2.2160e-2 (1.49e-3)} \\
  & $(10,10)$ & 2.1509e-2 (2.73e-3) $=$ & 1.3184e-1 (1.14e-2) $-$ & 2.9525e-2 (8.43e-3) $-$ & 2.3565e-2 (4.48e-3) $=$ & \textbf{2.1211e-2 (1.21e-3)} \\
  & $(20,10)$ & \textbf{1.6870e-2 (2.31e-3)} $+$ & 1.2212e-1 (1.21e-2) $-$ & 2.8370e-2 (7.65e-3) $-$ & 1.8648e-2 (3.84e-3) $=$ & 1.8533e-2 (1.18e-3) \\
\hline
DF9 & $(5,10)$ & 4.3703e-1 (3.03e-2) $-$ & \textbf{3.1328e-1 (1.60e-2)} $+$ & 4.8363e-1 (3.13e-2) $-$ & 7.7547e-1 (1.92e-1) $-$ & 4.0564e-1 (4.08e-2) \\
  & $(10,10)$ & 3.0413e-1 (3.59e-2) $-$ & 3.3210e-1 (3.37e-2) $-$ & 3.3862e-1 (3.93e-2) $-$ & 8.1361e-1 (1.99e-1) $-$ & \textbf{1.9262e-1 (1.87e-2)} \\
  & $(20,10)$ & 1.7902e-1 (2.22e-2) $-$ & 3.4970e-1 (2.57e-2) $-$ & 2.4225e-1 (4.08e-2) $-$ & 6.0341e-1 (9.43e-2) $-$ & \textbf{9.5476e-2 (1.07e-2)} \\
\hline
DF10 & $(5,10)$ & \textbf{9.7980e-2 (2.61e-3)} $+$ & 3.0834e-1 (3.87e-2) $-$ & 2.0123e-1 (2.18e-2) $-$ & 1.3874e-1 (7.73e-3) $+$ & 1.7749e-1 (1.38e-2) \\
  & $(10,10)$ & \textbf{1.1304e-1 (5.80e-3)} $+$ & 3.6355e-1 (4.95e-2) $-$ & 2.2071e-1 (2.93e-2) $-$ & 1.6228e-1 (1.12e-2) $+$ & 1.8651e-1 (1.29e-2) \\
  & $(20,10)$ & \textbf{7.9751e-2 (2.72e-3)} $+$ & 2.7100e-1 (3.61e-2) $-$ & 1.2332e-1 (1.13e-2) $+$ & 1.0991e-1 (7.38e-3) $+$ & 1.4482e-1 (8.94e-3) \\
\hline
DF11 & $(5,10)$ & 1.1908e-1 (4.21e-3) $-$ & 2.2049e-1 (2.79e-2) $-$ & 1.1856e-1 (9.44e-3) $-$ & 1.3943e-1 (4.35e-3) $-$ & \textbf{1.0610e-1 (2.58e-3)} \\
  & $(10,10)$ & 1.0819e-1 (3.96e-3) $-$ & 2.2451e-1 (2.86e-2) $-$ & 1.0668e-1 (7.32e-3) $=$ & 1.3088e-1 (8.10e-3) $-$ & \textbf{1.0413e-1 (3.39e-3)} \\
  & $(20,10)$ & \textbf{9.8137e-2 (3.21e-3)} $+$ & 2.2025e-1 (1.98e-2) $-$ & 1.0191e-1 (8.45e-3) $+$ & 1.3136e-1 (7.24e-3) $-$ & 1.0580e-1 (3.74e-3) \\
\hline
DF12 & $(5,10)$ & 4.2404e-1 (9.20e-2) $-$ & 6.8303e-1 (4.25e-2) $-$ & 3.1175e-1 (3.76e-2) $-$ & 5.7487e-1 (9.17e-2) $-$ & \textbf{2.6980e-1 (1.70e-2)} \\
  & $(10,10)$ & 3.3565e-1 (8.69e-2) $-$ & 6.6219e-1 (6.19e-2) $-$ & \textbf{1.8843e-1 (1.72e-2)} $+$ & 4.9101e-1 (1.05e-1) $-$ & 2.1015e-1 (1.43e-2) \\
  & $(20,10)$ & 1.8594e-1 (2.18e-2) $=$ & 5.1648e-1 (5.54e-2) $-$ & \textbf{1.5238e-1 (1.74e-2)} $+$ & 3.3111e-1 (5.68e-2) $-$ & 1.7931e-1 (1.38e-2) \\
\hline
DF13 & $(5,10)$ & 1.1831e+0 (1.93e-1) $-$ & 4.1660e-1 (1.98e-2) $-$ & 3.4340e-1 (6.08e-2) $-$ & 4.4345e-1 (6.33e-2) $-$ & \textbf{1.5359e-1 (4.16e-3)} \\
  & $(10,10)$ & 2.2601e-1 (1.08e-2) $-$ & 4.3024e-1 (2.79e-2) $-$ & 2.0870e-1 (5.12e-2) $-$ & 2.3890e-1 (3.09e-2) $-$ & \textbf{1.3632e-1 (3.43e-3)} \\
  & $(20,10)$ & 1.3698e-1 (5.12e-3) $-$ & 4.4221e-1 (1.77e-2) $-$ & 1.7346e-1 (4.98e-2) $-$ & 1.5081e-1 (4.85e-3) $-$ & \textbf{1.2966e-1 (4.49e-3)} \\
\hline
DF14 & $(5,10)$ & 1.1685e+0 (3.89e-1) $-$ & \textbf{3.0630e-1 (5.23e-2)} $+$ & 4.8918e-1 (1.13e-1) $=$ & 8.3037e-1 (2.75e-1) $-$ & 4.3898e-1 (7.11e-2) \\
  & $(10,10)$ & 1.0587e+0 (6.17e-1) $-$ & 3.5630e-1 (6.59e-2) $+$ & \textbf{3.4468e-1 (1.37e-1)} $+$ & 6.8440e-1 (1.53e-1) $-$ & 4.0903e-1 (7.96e-2) \\
  & $(20,10)$ & 6.4135e-1 (5.96e-1) $=$ & 4.8800e-1 (8.00e-2) $-$ & \textbf{2.3354e-1 (8.57e-2)} $+$ & 6.5465e-1 (1.42e-1) $-$ & 3.8871e-1 (7.20e-2) \\
\hline
$+/-/=$ &  & 6/30/6 & 3/39/0 & 6/34/2 & 3/36/3 & -- \\
Best/All &  & 7/42 & 2/42 & 4/42 & 0/42 & \textbf{29/42} \\
\hline
\multicolumn{7}{l}{\footnotesize $+$, $-$, and $=$ indicate that the corresponding algorithm performs significantly better than, worse than, or equivalently to SPSR-DMOEA, respectively.}\\
\multicolumn{7}{l}{\footnotesize The best mean value in each row is highlighted in bold.}\\
\end{tabularx}
\end{table*}
Table \ref{tab:migd} presents the MIGD results of SPSR-DMOEA and the
compared algorithms on the DF test suite. The best mean value in each
case is highlighted in bold. As shown in the table, SPSR-DMOEA obtains
the best results in 29 out of 42 test cases.Compared with the four competitors, SPSR-DMOEA shows strong competitiveness under different dynamic settings. It achieves strong performance across on all three settings of DF1--DF9, indicating that the proposed prediction strategy can effectively improve the quality of the population after environmental changes. 
However, SPSR-DMOEA still shows relatively weak performance on DF10, DF12 and DF14. One possible explanation is that, when dealing with three-objective optimization problems, the constructed prediction skeleton may fail to accurately characterize the intrinsic structural relationships of the predicted population, leading to a degraded quality of the reconstructed population.
\section{CONCLUSION}
This paper proposes a dynamic environmental change response mechanism based on special-point skeleton reconstruction for generating predicted populations. By exploiting the geometric relationships among the centroid, knee point, and extreme points, the proposed mechanism reconstructs the population after environmental changes. Experimental results on benchmark test problems demonstrate the effectiveness of the proposed mechanism. However, the structural information of populations has not been fully exploited in the current framework. In future work, more structural characteristics of populations will be explored to further improve the quality of predicted populations.
\bibliographystyle{IEEEtran}
\bibliography{reference}
\end{document}